\documentclass[runningheads]{llncs}
\usepackage[T1]{fontenc}
\usepackage{marvosym}
\usepackage{makecell}                 
\usepackage{booktabs}                 
\usepackage{graphicx}
\usepackage{multirow}
\usepackage{graphicx}
\usepackage{subcaption}
\usepackage{amssymb}
\usepackage[ruled,linesnumbered]{algorithm2e}
\begin{document}
\title{COVER: A Heuristic Greedy Adversarial Attack on Prompt-based Learning in Language Models\thanks{Supported by Guangdong Provincial Key-Area Research and Development Program (2022B0101010005), Qinghai Provincial Science and Technology Research Program (2021-QY-206), National Natural Science Foundation of China (62071201), and Guangdong Basic and Applied Basic Research Foundation (No.2022A1515010119). \\ 
The corresponding author is Qingliang Chen. }}
%
%

\author{Zihao Tan\inst{1}\and
Qingliang Chen \inst{1}\ \Letter \and
Wenbin Zhu \inst{1} \and Yongjian Huang\inst{2}}
\authorrunning{Tan et al.}
\titlerunning{COVER}
%

\institute{$^1$Department of Computer Science, Jinan University, Guangzhou 510632, China \\
$^2$Guangzhou Xuanyuan Research Institute Co., Ltd., Guangzhou 510006, China
\email{tzhtyson@stu2022.jnu.edu.cn, tpchen@jnu.edu.cn}}


\maketitle              
\begin{abstract}
Prompt-based learning has been proved to be an effective way in pre-trained language models (PLMs), especially in low-resource scenarios like few-shot settings. However, the trustworthiness of PLMs is of paramount significance and potential vulnerabilities have been shown in prompt-based templates that could mislead the predictions of language models, causing serious security concerns. In this paper, we will shed light on some vulnerabilities of PLMs, by proposing a prompt-based adversarial attack on manual templates in black box scenarios. First of all, we design character-level and word-level heuristic approaches to break manual templates separately. Then we present a greedy algorithm for the attack based on the above heuristic destructive approaches. Finally, we evaluate our approach with the classification tasks on three variants of BERT series models and eight datasets. And comprehensive experimental results justify the effectiveness of our approach in terms of attack success rate and attack speed.

\end{abstract}
%
%
\section{Introduction}
The introduction of pre-trained language models (PLMs) has greatly revolutionized natural language processing with pre-training+fine-tuning paradigm. However, such a paradigm suffers from some drawbacks of high computational resources and poor inference due to too little or unbalanced data during fine-tuning \cite{ref_5}. To tackle this problem, prompt-based learning has been proposed in recent years \cite{ref_8}, which can stimulate the potential of language models with less data and computational resources by designing templates and verbalizers.

As for the template designs, however, malicious design like adversarial attacks will mislead the model predictions. In general, adversarial attacks on PLMs are divided into white-box \cite{ref_11,ref_12} and black-box \cite{ref_10} ones. The former requires obtaining information about the parameters, gradients, and structure of the model. In the latter case, only the output distribution of the model is needed. Nonetheless, existing research on adversarial attacks on prompt-based learning mainly focuses on white-box scenarios, while there is little study on black-box ones, which can generate more serious security concerns in practices.

To address the deficit, we propose a \textbf{C}haracter-level and w\textbf{O}rd-le\textbf{V}el h\textbf{E}uristic g\textbf{R}eedy (\textbf{COVER}) approach in this paper. First, we design character-level and word-level heuristic destruction rules against the manual template, which act to corrupt the template before each model's prediction. Then, we introduce a greedy strategy in the attack phase. We conducted extensive experiments with three BERT series models on eight classification tasks, and the experimental results have justified the destructive power and attack speed of our proposed method. In summary, the contributions of the paper are as follows:
\begin{itemize}
\item We present the manual template black-box attack method in prompt-based learning, which is an attack scenario with significant practical implications, and almost no other works focus on it.
\item We design character-level and word-level heuristic manual template destruction rules that can work before each prediction, and furthermore with a greedy approach based on the above rules.
\item Experiments show that our attack method achieves high attack success rates and low number of queries on most of the classification task datasets.
\end{itemize}

\section{The Proposed Method}

\begin{figure}[!h]
    \centering
    \includegraphics[width=0.9\textwidth]{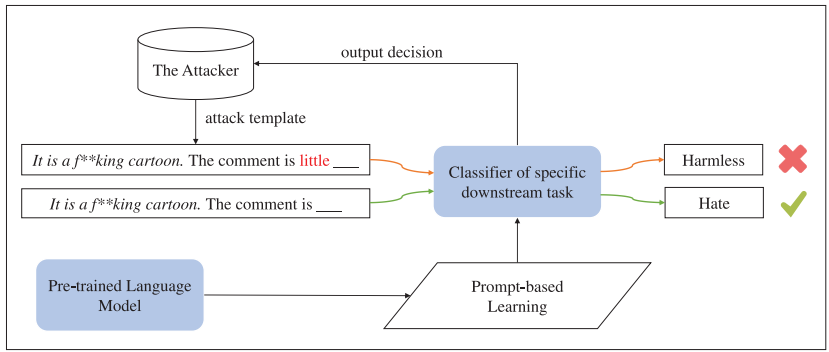}
    \caption{Overview of the adversarial attack in black-box scenarios}
    \label{fig:blackbox}
\end{figure}

Consider a publicly released PLM $f:X \to Y$ after few-shot tuning of a text classification task. An input text $x \in X$ is transformed by a clean template $T_c$ like $x'_c=T_c(x)$. Then it can be passed into the $f$ to make a correct prediction:
\begin{equation}
\mathop{\arg\max_{y_i \in Y}}\, P(y_i|x'_c)=y_{true}.
\end{equation}
where $y_{true}$ is the correct label. Attackers try to use a series of destruction rules to attack the clean template, fooling the PLM with the processed poisoned template $x'_p=T_p(x)$. And the classifier will finally predict wrongly:
\begin{equation}
\mathop{\arg\max_{y_i \in Y}}\, P(y_i|x'_p)\neq y_{true}.
\end{equation}
In our setting, it is worth noting that our attack scenario is totally based on the black-box ones, without the need of the gradient, score, structure and parameter information of the PLM to carry out the attack. The overview of prompt-based learning adversarial attack in black-box scenarios is shown in Figure \ref{fig:blackbox}.

Inspired by Chen et al. \cite{ref_16} for real-world attackers' sabotage rules on texts, we devise a series of character-level and word-level heuristic destruction rules of prompt-based learning. The difference is that they simply use six of the destriction rules (rule 1-5, 10) and did not normalize destruction level. Table~\ref{desrule} gives a summary of the whole destruction rules. 

\begin{table*}
\small
\centering

\caption{Destruction rules on template based on character-level and word-level}\label{desrule}
\resizebox{0.9\linewidth}{!}{
\begin{tabular}{c|c|l|l}
\Xhline{1pt}
Level & Rule & Description & Example\\
\hline
\multirow{6}*{Char}& (1) & Insert a space into words & $x$. The sen timent is <mask>\\
~& (2) & Insert a punctuation into words & $x$. The sent*iment is <mask>\\
~& (3) & Swap two adjacent character & $x$. The senitment is <mask>\\
~& (4) & Delete a character of words & $x$. The seniment is <mask>\\
~& (5) & Replace a character of words & $x$. The 5entiment is <mask>\\
~& (6) & Duplicate a character of words & $x$. The senttiment is <mask>\\
\hline
\multirow{4}*{Word}& (7) &  Exchange mask token's position & $x$. The <mask> sentiment is \\
~& (8) & Swap two word except mask token & $x$. The is sentiment <mask>\\
~& (9) & Add negative word after linking verb & $x$. The sentiment is little <mask>\\
~& (10) & Add prefixes and suffixes  & $x$. sad The sentiment is <mask> sad\\
\Xhline{1pt}
\end{tabular}}
\end{table*}

Now, we introduce a greedy attack strategy based on the heuristic destruction rules described above. Specifically, we use an ordered dictionary $Dict$ in the data structure, which includes the $value$ to record the time of successful attacks and the corresponding template is recorded by its $key$. The dictionary is arranged in descending order by values. For future data, the first $k$ templates with the top dictionary sorting are taken out as the candidate template set $C_{template}$:
\begin{equation}
C_{template} = \mathop{topk} \limits_{d_i \in Dict}(d_i.value)
\end{equation}

The complete algorithm pipeline is shown in Algorithm \ref{algorithm1}.

\begin{algorithm}[h]

\caption{Adversarial Attack by COVER}
\label{algorithm1}
\small
\KwIn{Text $x \in X$ with correct prediction $y_{true} \in Y$, clean template $T_c$, ordered dictionary $Dict$, destruction function $g_i$, PLM $f$, iteration $ITER$, repeat time $REP$, and max length $LEN$.}
\KwOut{Attack success (true) or fail (false)}
$iter\leftarrow 0$\;
\If{$len(Dict)>0$}{$C_{template} = Dict.get\_top\_k()$\;
\For{$t_p$ in $C_{template}$}{\If{$f(t_p(x)) \neq y_{true}$}{$Dict.record(t_p)$\;$iter \leftarrow iter+1 $\;$return$ true\;}}}

$T_p\leftarrow g_9(g_{10}(T_c))$\;
$iter \leftarrow iter+1 $\;
\If{$f(T_p(x)) \neq y_{true}$}{$Dict.add(T_p)$\;
$return$ true\;}
\While{$iter<ITER\cdot REP$ and $len(T_c(x))<LEN$}{$i \leftarrow Random(rules), rules \in [1,9]$\; $T_p=g_i(T_p)$\;\If{$f(T_p(x)) \neq y_{true}$}{$Dict.record(T_p)$\;$iter \leftarrow iter+1 $\;$return$ true\;}}
$return$ false\;
\end{algorithm}

\section{Experiments}

\subsubsection{Dataset and Victim Model}
\begin{table}
\centering
\caption{Dataset details}\label{dataset}
\resizebox{0.9\linewidth}{!}{
\begin{tabular}{c|c|c|l}
\Xhline{1pt}
Region & Dataset & Class & Description\\
\hline
\multirow{2}*{Sentiment} & SST2 & 2 & Movie reviews and human comments data\\
~& IMDB & 2 & Large movie review dataset\\
\hline
\multirow{2}*{Disinformation}& Amazon-LB &  2 & Small subsets of Amazon Luxury Beauty Review\\
~& CGFake & 2 & Computer-generated Fake Review Dataset\\
\hline
\multirow{2}*{Toxic}& HSOL &  2 & Hate offensive speech dataset\\
~& Jigsaw2018 & 2 & Toxic Comment Classification Challenge in Kaggle\\
\hline
\multirow{2}*{Spam}& Enron &  2 & Collections of emails include legitimate and spam\\
~& SpamAssassin & 2 & Collections of emails include ham and spam\\
\Xhline{1pt}
\end{tabular}}
\end{table}

The datasets we chose to evaluate our method have four domains as shown in Table \ref{dataset}. The sentiment domain includes SST2 \cite{ref_22} and IMDB \cite{ref_23}, and the remaining disinformation, toxic and spam domains consist of six datasets which are compiled by Chen et al. \cite{ref_16}. And we use three pre-trained language models of the BERT family: BERT-base (109M) \cite{ref_19}, RoBERTa-base (125M) and RoBERTa-large (355M) \cite{ref_20}.

\subsubsection{Parameter Settings}

For each dataset in the pre-trained model, 8 shots of few-shot tuning were performed. We designed two sets of manual templates for each dataset separately, each containing two and swapped sentences of the original input. On the training phase, we tune 10 epochs by AdamW optimizer \cite{ref_21} with learning rate of $1e-5$ and weight decay $1e-2$. On the attack phase, we iterate 30 times for each sentence and the $k$ value of the ordered dictionary is set to 2.

\subsubsection{Metrics and Baselines}
We apply two evaluation metrics: (1) Attack success rate (ASR): the percentage of data which has been attacked successfully. (2) Attack efficiency (Query): the query times to the PLM after crafting a victim input.

Since there is no prior work for black-box adversarial attacks on prompt-base learning, we think about two baselines. The first is the heuristic attack method ROCKET proposed by Chen \cite{ref_16} et al. for text to templates, we keep the stop words with minor modifications and name it \textbf{rocket-prompt}. And the other baseline is character-level and word-level heuristic approaches without greedy strategy and is labelled \textbf{COVE}.

\subsubsection{Experimental Results}
\begin{table}
    \centering
    \caption{COVER versus rocket-prompt and COVE in ASR and Query. }
    \label{mainresult}
    \resizebox{0.9\linewidth}{!}{
    \begin{tabular}{c|c|cccccccc}
        \Xhline{1pt}
        \multicolumn{2}{c}{Task} &\multicolumn{2}{c}{Sentiment} & \multicolumn{2}{c}{Disinformation} & \multicolumn{2}{c}{Toxic} & \multicolumn{2}{c}{Spam} \\
        \hline
        \multirow{2}{*}{PLM} & \multirow{2}{*}{Method|Dataset} & \multicolumn{2}{c}{\underline{SST2}} & \multicolumn{2}{c}{\underline{Amazon-LB}} & \multicolumn{2}{c}{\underline{HSOL}} & \multicolumn{2}{c}{\underline{Enron}} \\
        & & ASR(\%) & Query & ASR(\%) & Query & ASR(\%) & Query & ASR(\%) & Query \\
        \hline
        \multirow{3}{*}{BERT-base} & rocket-prompt & 94.8 & 3127.5 & \textbf{100} & 1537.8 & 14.4 & 13118.5 & 58.3 & 6985.3 \\
        & COVE & 99.8 & 962 & \textbf{100} & 1006.3 & 57.2 & 6638.8 & 85.8 & 2668 \\
        & COVER & \textbf{100} & \textbf{494} & \textbf{100} & \textbf{773} & \textbf{89.9} & \textbf{2058} & \textbf{96.7} & \textbf{1008.3} \\
        \hline
        \multirow{3}{*}{RoBERTa-base} & rocket-prompt & 92.3 & 4180.5 & 81.5 & 5535.8 & 22.6 & 12036.3 & 94.3 & 2414.5 \\
        & COVE & 97.5 & 1698.8 & 90.2 & 3222.8 & 35.3 & 6638.8 & 97.2 & 1402 \\
        & COVER & \textbf{99.9} & \textbf{757.5} & \textbf{98} & \textbf{1527} & \textbf{87.5} & \textbf{2293.3} & \textbf{99.4} & \textbf{998} \\
        \hline
        \multirow{3}{*}{RoBERTa-large} & rocket-prompt & 92.3 & 4021.5 & 93 & 4417.75 & 3.7 & 14560 & 80.4 & 4962.8 \\
        & COVE & 97.3 & 1663.5 & 93.5 & 4136.25 & 13.2 & 12456.8 & 90.9 & 2621.5 \\
        & COVER & \textbf{99.8} & \textbf{733.25} & \textbf{95.8} & \textbf{3150} & \textbf{27.7} & \textbf{10624} & \textbf{93.8} & \textbf{1933} \\
        \hline
        \multicolumn{2}{c}{Average Accuracy (\%)} & \multicolumn{2}{c}{83} & \multicolumn{2}{c}{71.8} & \multicolumn{2}{c}{70.5} & \multicolumn{2}{c}{76.6} \\
         \hline
         
        \multicolumn{10}{c}{}\\
         \hline
          \multirow{2}{*}{PLM} & \multirow{2}{*}{Method|Dataset} & \multicolumn{2}{c}{\underline{IMDB}} & \multicolumn{2}{c}{\underline{CGFake}} & \multicolumn{2}{c}{\underline{Jigsaw2018}} & \multicolumn{2}{c}{\underline{SpamAssassin}} \\
        & & ASR(\%) & Query & ASR(\%) & Query & ASR(\%) & Query & ASR(\%) & Query \\
        \hline
        
        \multirow{3}{*}{BERT-base} & rocket-prompt & 24.1 & 12624.5 & 99.9 & 1537.8 & 20.1 & 12392 & 74.5 & 4699 \\
        & COVE & 72.8 & 5834.8 & \textbf{100} & 1097.5 & 46.2 & 7989 & 92.3 & 1862.3 \\
        & COVER & \textbf{94.9} & \textbf{1703} & \textbf{100} & \textbf{674.3} & \textbf{86.4} & \textbf{2473.5} & \textbf{99.8} & \textbf{536.75} \\
        \hline
        
        \multirow{3}{*}{RoBERTa-base} & rocket-prompt & 40.4 & 10479.3 & 97.5 & 3417.8 & 28.5 & 11673.3 & 99.7 & 1725.3 \\
        & COVE & 64.6 & 6900.8 & 97.4 & 2798.8 & 37.6 & 9307 & 99.8 & 984.5 \\
        & COVER & \textbf{92.3} & \textbf{2250.8} & \textbf{98.5} & \textbf{1589.3} & \textbf{77.2} & \textbf{3776} & \textbf{100} & \textbf{525.8} \\
        \hline
        
        \multirow{3}{*}{RoBERTa-large} & rocket-prompt & 49.1 & 10888.5 & \textbf{95.2} & 4828.25 & 15.9 & 13328.8 & 93.3 & 4043.5 \\
        & COVE & 79.1 & 5719 & 91.1 & 4683 & 22.2 & 11376.8 & 96.2 & 2520 \\
        & COVER & \textbf{96.1} & \textbf{1607.3} & 91.8 & \textbf{4369.5} & \textbf{31.7} & \textbf{10225} & \textbf{97.4} & \textbf{1994.25} \\
        \hline
        \multicolumn{2}{c}{Average Accuracy (\%)} & \multicolumn{2}{c}{79.2} & \multicolumn{2}{c}{62.5} & \multicolumn{2}{c}{74.2} & \multicolumn{2}{c}{81.8} \\
        
        \Xhline{1pt}
    \end{tabular}}
\end{table}

Table \ref{mainresult} shows the performance of our proposed COVER. The ASR of COVER achieves an average accuracy of 96\%, 94.1\% and 75.3\% in BERT-base, RoBERTa-base and RoBERTa-large, respectively, significantly outperforming that of rocket-prompt and COVE. And COVER has the least Query times in all cases, where it is almost one-sixth of that of rocket-prompt and almost one-third to one-half of that of COVE in the Sentiment domain. This demonstrates the vulnerabilities of prompt-based learning where an attacker can corrupt PLM predictions through heuristic greedy means, which needs to be taken into account by real-world practitioners.

\section{Conclusion}
In this paper, we explore black-box attacks for prompt-based learning, which carries more practical values. First, we design a series of heuristic template destruction rules at character-level and word-level. Then we propose a greedy strategy based on this to mimic real-world malicious attacks. And finally the experimental results justify the power of our approach in terms of both attack success rate and speed, exhibiting great vulnerability in prompt-based learning.

%
%
%
%

\end{document}